\documentclass{article}
\usepackage{spconf,amsmath,graphicx,arydshln,cite,balance,url}
\newcommand{\chapternote}[1]{{%
  \renewcommand{\thefootnote}{\fnsymbol{footnote}}%
  \footnotetext[0]{#1}
}}

\title{A DEEP LEARNING APPROACH TO OBJECT AFFORDANCE SEGMENTATION}
\name{Spyridon Thermos $^{1,2}$ \qquad Petros Daras $^{2}$ \qquad Gerasimos Potamianos $^{1}$}
\address{$^{1}$ Department of Electrical and Computer Engineering, University of Thessaly, Volos, Greece\\
$^{2}$ Information Technologies Institute, Centre for Research and Technology Hellas, Thessaloniki, Greece\\
{\tt\small spthermo@e-ce.uth.gr} \qquad {\tt\small daras@iti.gr} \qquad {\tt\small gpotam@ieee.org}
}

\begin{document}
\ninept
\maketitle
\begin{abstract}
Learning to understand and infer object functionalities is an important step towards robust visual intelligence. Significant research efforts have recently focused on segmenting the object parts that enable specific types of human-object interaction, the so-called ``object affordances". However, most works treat it as a static semantic segmentation problem, focusing solely on object appearance and relying on strong supervision and object detection. In this paper, we propose a novel approach that exploits the spatio-temporal nature of human-object interaction for affordance segmentation. In particular, we design an autoencoder that is trained using ground-truth labels of only the last frame of the sequence, and is able to infer pixel-wise affordance labels in both videos and static images. Our model surpasses the need for object labels and bounding boxes by using a soft-attention mechanism that enables the implicit localization of the interaction hotspot. For evaluation purposes, we introduce the SOR3D-AFF corpus, which consists of human-object interaction sequences and supports 9 types of affordances in terms of pixel-wise annotation, covering typical manipulations of tool-like objects. We show that our model achieves competitive results compared to strongly supervised methods on SOR3D-AFF, while being able to predict affordances for similar unseen objects in two affordance image-only datasets.
\end{abstract}
\chapternote{Supported by the EC under contract H2020-820742 HR-Recycler.}
\section{Introduction}
\label{sec:intro}
Recent visual understanding systems are capable of detecting and recognizing objects in 2D/3D scenes~\cite{ssd, 7780460, 7472087, 8237584, 8417976}. However, a truly intelligent system should not only recognize an object, but also understand its functionality.  Gibson~\cite{gibson1} provides a way to reason about object functionalities and defines them as affordances, namely the types of actions that humans typically perform when interacting with them. Understanding object affordances and being able to localize and segment them is an important step towards robust scene understanding and active embodiment~\cite{lyubova, jayaraman}.

\begin{figure}[t]
    \centering
    \includegraphics[width=0.95\columnwidth]{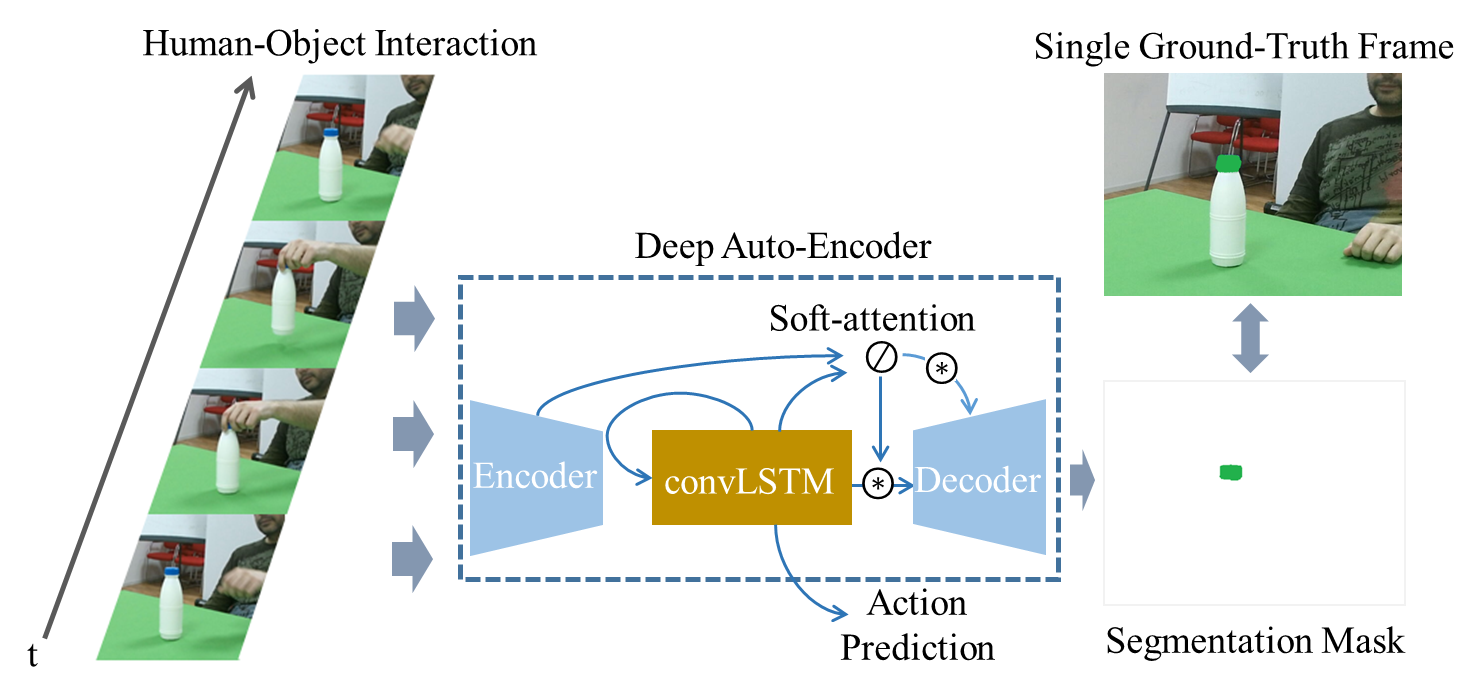}
    \vspace{-0.1in}
    \caption{Overview of the proposed approach. Our deep autoencoder processes frames of human-object interaction sequences and infers pixel-wise affordance label predictions. A soft-attention mechanism at the latent space performs implicit localization of the interaction hotspot. The model is able to infer affordances both in videos and static images. In this example, the top part of the bottle is segmented and associated with the ``lift" affordance label.}
    \label{fig:overview}
    \vspace{-0.15in}
\end{figure}

Object affordance segmentation, \textit{i.e.}~the pixel-wise identification of the object part that enables a specific interaction, is a challenging task that has been mostly treated as a static semantic segmentation problem, usually coupled with object detection. For example, Myers~\textit{et al.}~\cite{7139369} use hierarchical matching pursuit, as well as normal and curvature features derived from RGB-D data, to learn pixel-wise labeling of affordances for common household objects, while Nguyen~\textit{et al.}~\cite{7759429} propose an encoder-decoder architecture to predict pixel-wise affordances based on depthmaps. Do~\textit{et al.}~\cite{8460902} expand the architecture of \cite{7759429} by adding a region proposal network (RPN)~\cite{7485869} to predict the bounding box of the target object and also investigate the joint learning of detecting and segmenting the object affordance part. All aforementioned works rely on strong supervision, as each object affordance part must be fully annotated at pixel-level. On the other hand, Sawatzky~\textit{et al.}~\cite{8100035} propose a weakly-supervised setting using CNNs and keypoints annotation to predict reasonable but not precise pixel-level labels, which are then refined using the GrabCut algorithm~\cite{1015720}. Although these methods report satisfactory results, they are dependent on successful object detection and the absence of occlusions by the interacting hand.

In addition to the above, there are affordance recognition approaches that utilize information related to the interaction such as the estimated human and hand poses, yet in a static manner~\cite{kragic2, Yao_2013_ICCV}. In the affordance reasoning domain though, this interaction-related information is used in a ``learning from observation" perspective, exploiting the spatio-temporal nature of the interaction. In particular, Fang~\textit{et al.}~\cite{demo2vec2018cvpr} present ``Demo2Vec" that learns spatio-temporal embeddings from product demonstrations and predicts keypoints on the object affordance part. Similarly, Nagarajan~\textit{et al.}~\cite{Nagarajan2018GroundedHI} propose a model that infers spatial hotspot maps on static images using gradient-weighted attention maps for pre-defined actions. However, these methods focus on predicting heatmaps on target object images and not on the frames of the processed sequences. 

In this paper, we surpass the aforementioned limitations, while adopting the ``learning from observation" scheme for the object affordance segmentation task. For this purpose, we propose a deep spatio-temporal autoencoder that learns from human-object interaction videos, combining appearance and motion information at the encoder and predicting pixel-level affordance labels at the decoder. Inspired by previous work~\cite{Zhang_2018_CVPR, Liu2019BraidNetBS} we design a soft-attention module, placed at the bottleneck, which is responsible for the implicit localization of the hand-object interaction. This mechanism fuses the frame-level spatial information with the video-level temporal one, forcing the network to focus on the object part that participates in the interaction. The spatio-temporal feature of the latent space is also used for action prediction, which is highly correlated with the object affordance. The proposed model exploits both RGB and depth representations, which are further used to compute the 3D optical flow of the interactions. Note that during inference, our model is able to infer pixel-wise affordance labels both on videos and static images. Additionally, we introduce the SOR3D-AFF\footnote{Available at \url{http://sor3d.vcl.iti.gr/}} dataset that consists of 1201 human-object interaction sequences and supports 9 affordance types of common household objects. We utilize SOR3D-AFF to evaluate our model performance over the state-of-the-art, while using the UMD~\cite{7139369} and IIT-AFF~\cite{8206484} datasets for qualitative evaluation. An overview of our approach is depicted in Fig.~\ref{fig:overview}.

The remainder of the paper is organized as follows: Section 2 details the available input representations, the architecture of the proposed deep autoencoder, and the soft-attention mechanism,  Section 3 provides a description of SOR3D-AFF, while Section 4 presents our experiments. Finally, Section 5 summarizes the paper.

\section{Approach}
\label{sec:approach}
Our approach aims to exploit the spatio-temporal nature of human-object interaction. An end-to-end autoencoder, trained using sequences of RGB-D data, learns to predict pixel-wise affordance labels, while an attention mechanism placed at the latent space of the model is responsible for implicitly localizing the interaction hotspot. Note, that the proposed autoencoder is able to infer affordance predictions on both video and static image data.

\vspace{-0.1in}
\subsection{Input Streams}
\label{sec:loss}
\begin{figure}[t]
    \centering
    \includegraphics[width=.9\columnwidth]{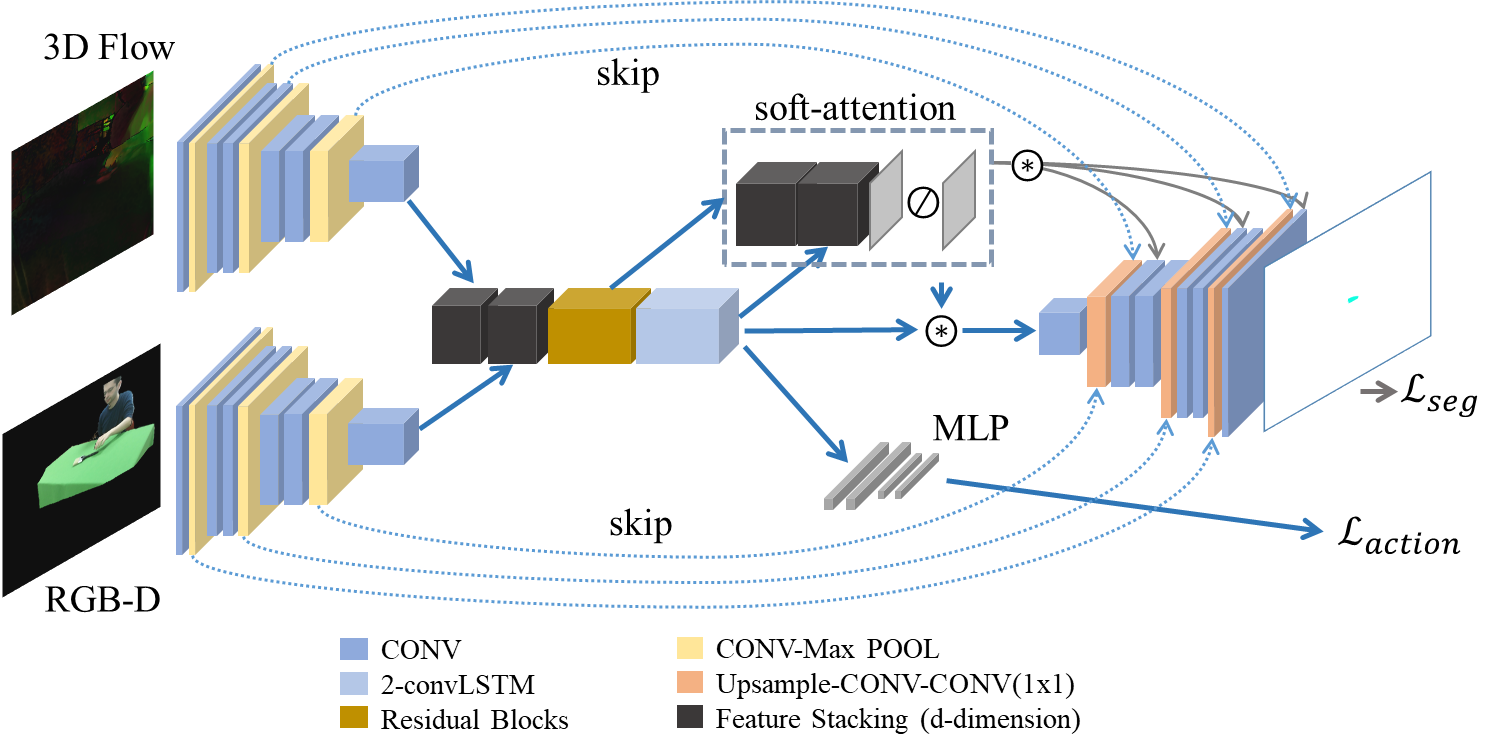}
    \vspace{-0.2in}
    \caption{Detailed model architecture of the proposed deep autoencoder. From left to right: a) the model receives RGB-D and 3D flow information using two convolutional encoders and fuses the encoded features, b) the latent space comprises one residual block and two convolutional LSTMs, followed by a soft-attention mechanism (dashed line), c) a decoder receives the output of the attention module to predict a segmentation map, and d) a fully connected network receives the convolutional LSTM output to predict the action class. The skip connections enable direct activations propagation from the encoder to the decoder.}
    \label{fig:net}
    \vspace{-0.2in}
\end{figure}

As depicted in Fig.~\ref{fig:net} and detailed in Section~\ref{sec:model}, the model encoder consists of two streams, one for RGB-D and the second for motion information processing. Intuitively, we use the former to force the model to learn appearance representations, while the later encodes the hand movement during the interaction.

We choose to combine RGB and depth information by stacking the color image and the depthmap along the channel dimension forming a $4\times H\times W$ input, where $H$ and $W$ represent the height and the width of the input image/depthmap. Further, we use the 3D optical flow of the sequence to represent the motion information. In particular, we utilize the algorithm proposed in~\cite{jaimez}, which computes the 3D motion vectors between two pairs of RGB-D images, and  colorize them by normalizing each axis values within $[0, 255]$, thus transforming them into a three-channel image of size $3\times H\times W$. This colorization enables the exploitation of transfer learning by using deep learning models pre-trained on large-scale image datasets.

\vspace{-0.1in}
\subsection{Deep Autoencoder Architecture}
\label{sec:model}
Our data-driven approach is realized as the deep autoencoder depicted in Fig.~\ref{fig:net}. Its structure is inspired by the U-Net architecture~\cite{ronn} and consists of an encoder, a latent part, and a decoder.

The encoder follows the typical structure of a VGG CNN~\cite{vgg} and consists of 11 convolutional (CONV) layers each followed by a Rectified Linear Unit (ReLU) activation function. The input is downsampled 3 times prior to the latent space using max pooling layers with $2\times2$ kernels and stride 2. Since we use RGB-D and 3D optical flow information, we utilize two identical streams for the encoder, which are fused prior to the latent part of the model. Let $X^{d\times h\times w}_{RGBD}$ be the feature of the RGB-D encoder stream, and $X^{d\times h\times w}_{Flow}$ be the corresponding 3D optical flow one, where $d=512$ is the number of channels, and $h=H/8$, $w=W/8$ are the height and width of both features after the 3 downsampling layers. Then, the two features are stacked along the channel dimension and are convolved with $d$ kernels of $1\times 1$ size. This step produces the final encoder activation map, $\Tilde{X}^{d\times h\times w}_{Enc}$. 

The latent part consists of a residual block and of 2 convolutional LSTM (convLSTM) layers. The residual block follows the ReLU-CONV-ReLU-CONV structure, adopting the pre-activation method and the identity mapping proposed in~\cite{he} for performance improvement. The latent part is followed by a soft-attention mechanism, detailed in the next section, which fuses the activations after the residual block with the second convLSTM output.

The decoder shares similar structure with the encoder, consisting of 14 CONV-ReLU layers. Prior to segmentation prediction, the spatio-temporal feature of the latent part is upsampled 3 times using nearest neighbor interpolation coupled with a CONV layer. Note that each max pooling layer is connected with the respective upsampling layer using a skip connection. Subsequently, the activations after upsampling are concatenated with the ones from the corresponding skip connection. After concatenation, a CONV layer with $1\times1$ kernel size follows, forcing intra-channel correlation learning. The output feature of the decoder is a $C\times H\times W$, where $C$ denotes the number of affordance classes.

All aforementioned parts of the model are fully convolutional. However, we utilize a simple multi-layer perceptron (MLP) to classify the performed action in parallel with the segmentation process. The MLP receives the output of the second convLSTM and consists of 3 fully connected layers.

The proposed model is trained using a spatio-temporal loss computed at the last frame of the sequence, denoted as $\mathcal{L}_{total} = \lambda_{1} \mathcal{L}_{seg} + \lambda_{2} \mathcal{L}_{action}$, where $\lambda_{1}, \lambda_{2} \in [0,1]$ are hyperparameters that add to 1, $\mathcal{L}_{seg}$ is the per pixel cross entropy of the predicted and ground truth affordance labels, and $\mathcal{L}_{action}$ is the cross entropy of the predicted and ground truth action labels. Note that we normalize $\mathcal{L}_{seg}$ over the number of pixels, and use $\mathcal{L}_{action}$ to force the model to understand the perceived interaction, thus helping the soft-attention mechanism to localize the interaction hotspot.

\subsection{Soft-Attention Mechanism}
\label{sec:attention}
Since the actual affordance part of the object is only a small portion of the input scene, we choose to design a soft-attention mechanism, which learns to focus on the hotspot of the human-object interaction. 

Using detection mechanisms, such as the widely-used RPN, to localize the object before predicting its affordance requires extra knowledge about its class label and bounding box. Besides being costly to acquire, this extra knowledge adds significant complexity to the model architecture and does not contribute to the generalization of the method to unseen objects. 

Instead, we propose an object-agnostic approach that forces the model to focus on the interaction hotspot based of the spatio-temporal information of the processed sequence. Our soft-attention mechanism is implemented in 3 steps. First, let $X^{d\times h\times w}$ be the spatial feature after the residual block of the latent space, and $\Bar{X}^{d\times h\times w}$ the spatio-temporal one after the second convLSTM. We then stack the two activation maps at the channel dimension and convolve the produced feature using a kernel of $1\times 1$ size. Second, we use the softmax function to normalize the activation values to the $[0,1]$ space, forming the ``excitation" mask $M^{1\times h\times w}$. Finally, $M$ is multiplied with each channel of $\Bar{X}$ in an element-wise manner and then upsampled and re-applied to the activation maps after each upsampling layer of the decoder. The multi-layer masking forces the model to focus on the interaction hotspot at different levels of granularity. The soft-attention mechanism is also visualized in Fig.~\ref{fig:net} (dashed line).

\begin{figure}[t]
    \centering
    \includegraphics[width=.95\columnwidth]{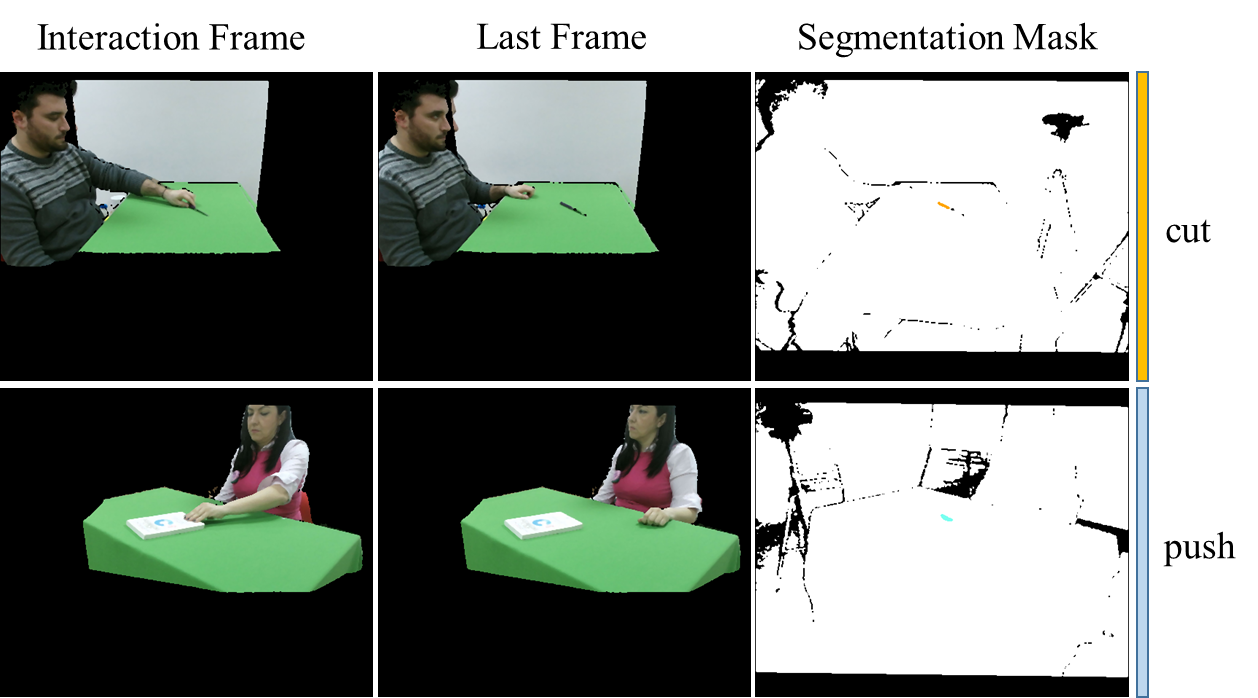}
    \vspace{-0.1in}
    \caption{Two indicative processed SOR3D-AFF samples. The 1st column depicts a frame sampled during the human-object interaction, while the 2nd one depicts the last frame of the sequence with the object at rest. Only the last frame is annotated as shown in the 3rd column. Each affordance label is visualized in different color as indicated by the colorized bars on the right.}
    \label{fig:dataset}
    \vspace{-0.2in}
\end{figure}

\vspace{-0.05in}
\section{Datasets}
\label{sec:datasets}
\vspace{-0.05in}
The main goal of our paper is to develop a model that learns to segment the object affordance part based on human-object interaction sequences. In order to train our model and provide a benchmark for other approaches, we need a dataset with subjects interacting with various objects. For this purpose we created a subset of the SOR3D sensorimotor dataset~\cite{8099496}, denoted as SOR3D-AFF. Its details and a brief description of UMD and IIT-AFF that are used for the qualitative evaluation follow.

\noindent \textbf{SOR3D-AFF:} The dataset consists of 1201 RGB-D interaction sequences, each pixel-wise annotated only at the last frame. It supports 9 affordance types, namely ``grasp", ``cut", ``lift", ``push", ``rotate", ``hammer", ``squeeze", ``paint", and ``type", of 10 common household objects, such as ``pitcher" and ``knife". We split the dataset into a training and a validation set, consisting of 962 and 239 interaction sequences, respectively. Some indicative annotated samples are depicted in Fig.~\ref{fig:dataset}. The RGB and depthmap frames pixel resolution is $1920\times 1080$ and $512\times 424$, respectively. However, we choose to map each RGB frame to the corresponding depthmap resolution, as visualized in Fig.~\ref{fig:dataset}, in order to jointly use the streams. Besides affordance pixel-wise annotation, each sequence is further annotated with an action label complementary to the corresponding affordance, \textit{i.e.}~``grasping", ``squeezing". 

\noindent \textbf{UMD and IIT-AFF:} UMD provides pixel-level affordance labels for 105 kitchen, workshop, and garden tools. The tools were collected from 17 different categories covering 7 affordances, namely ``grasp", ``cut", ``scoop", ``contain", ``pound", ``support", ``wrap-grasp". Both RGB and depthmap frames are in $640\times 480$ pixel resolution. IIT-AFF consists of a combination of images from ImageNet~\cite{5206848} and a collection from 2 RGB-D sensors at various pixel resolutions. All images depict cluttered scenes with multiple objects. The dataset supports 9 affordance classes, namely ``contain", ``cut", ``display", ``engine", ``grasp", ``hit", ``pound", ``support", ``w-grasp", and provides pixel-wise affordance annotations and bounding boxes.

\vspace{-0.05in}
\section{Experiments}
\label{sec:experiments}
\vspace{-0.05in}
In this section, we present the quantitative and qualitative evaluations of the proposed model, as well as an ablation study to demonstrate each individual model component contribution. 

\vspace{-0.1in}
\subsection{Implementation Details}
\label{quant}
All images and video frames utilized as inputs to the model are resized to $300\times 300$ pixel resolution, while each video is subsampled to 10 FPS. We pre-train both encoders on separate datasets; the RGB-D encoder coupled with an LSTM is trained for 50 epochs on the UTKinect action recognition dataset~\cite{xia2012view}, while for the colorized 3D flow encoder the weights of a VGG16 trained on ImageNet are used. The rest of the model weights are initialized using Xavier initialization~\cite{glorot10a}. The model is fine-tuned for 200 epochs, using batch size equal to $2$, Adam optimization~\cite{1412}, and learning rate set to $2\times 10^{-5}$. Further, we set $\lambda_{1} = 0.2$ and $\lambda_{2} = 0.8$ for the first 150 epochs, as action recognition is a critical step towards affordance segmentation and should converge faster that the total loss. For the last 50 epochs, both hyperparameters are tuned to $0.5$. During inference our model is able to predict a segmentation mask in 22ms.

We choose to quantitatively compare our model with AffordanceNet~\cite{8460902}, a convolutional autoencoder that utilizes an RPN in order to restrict affordance segmentation to a detected bounding box. We re-implement AffordanceNet to receive $300\times 300$ inputs and train it for 50 epochs with the batch size set to $8$ and learning rate equal to $2\times 10^{-5}$. All models are implemented in PyTorch~\cite{pytorch} and trained on an Nvidia Titan X GPU.

\vspace{-0.1in}
\subsection{Quantitative Evaluation}
\label{ablation}
We use 2 different metrics to assess model performance: a) the Intersection over Union (IoU), and b) the F$_{\beta}$-score measure for $\beta = 1$. IoU quantifies the overlap between the predicted and target affordance mask, while F-score provides helpful insight about the model robustness based on  false positive and negative predictions. Since some affordances are associated with more objects, we choose to evaluate the performance of the model using a variation of F-score, namely the weighted F-score denoted as F$_{\beta}^{w}$. Since ``grasp" and ``lift" are the most dominant affordance labels, we set their weight to $0.2$, while for the next dominant label ``push" we set it to $0.1$. The remaining weights are set to $0.083$ so that they all sum to 1.

Table~\ref{table:res} reports the overall performance of the proposed autoencoder on the SOR3D-AFF test set. Since there is no equivalent model for inferring pixel-wise affordance labels based on videos, only our model results are reported (top row). For the static image-based affordance prediction, our model is compared to the AffordanceNet and achieves competitive results (bottom rows). Note, that the goal of this work is to perform affordance segmentation using weak supervision, \textit{i.e.}~ground truth only for the last frame, and without exploiting additional annotations, such as object class and bounding box. The results also support our argument that a model can be trained using interaction sequences and infer affordance labels for both videos and static images.

We also present results per affordance category in Table~\ref{table:aff}, based on both video (top) and static image (bottom) inference. From the reported results, we can observe the superiority of the dominant affordances, \textit{i.e.}~these associated with most of the objects, such as ``grasp" and ``lift", as well as the adequate performance of complex affordances that change the visual representation of the object, such as ``rotate". Note, that affordance label weighting leads to a slightly better overall performance in terms of F-score, which is expected given the very confident predictions for the dominant affordances.

In order to demonstrate the contribution of each individual component to the proposed model architecture, we perform an ablation study using the following variations: a) single-stream RGB-only encoder, b) single-stream RGB-only encoder and soft-attention mechanism, and c) two-stream encoder for RGB-only and 2D optical flow, and soft-attention mechanism. Note, that the same model variations are investigated for the RGB-D and 3D optical flow information.

Table~\ref{table:abl} (top) presents the results of the aforementioned variations using the full sequences of the SOR3D-AFF test set to infer the affordance part of the object. The performance of the best RGB and RGB-D models when inferring affordances on static images can be visualized at the bottom part of the same table. Evidently, RGB-D information leads to better overall performance, while the integration of the soft-attention mechanism leads to the greatest relative improvement for both RGB and RGB-D based models, \textit{i.e.}~5.48\% (F$_{\beta}$) and 9.38\% (IoU) respectively. Finally, we observe that the use of a second stream that processes optical flow information leads to further improvement of the overall model performance, mostly due to its contribution to the action classification part of the network.

\begin{table}[t]
\caption{Overall object affordance segmentation results on the SOR3D-AFF test set based on video (top) and static image inference (bottom).}
\vspace{-0.1in}
\normalsize
\begin{center}
\begin{tabular}{l c c c}
\hline
\\[-0.95em]
\textbf{Model} & IoU & F$_{\beta}$ & F$_{\beta}^{w}$\\
\hline
\\[-0.85em]
Ours (RGB-D, attention, 3Dflow) & 0.72 & 0.80 & 0.81 \\
\hdashline
\\[-0.85em]
Ours (RGB-D, attention, 3Dflow) & 0.54 & 0.58 & 0.59 \\
AffordanceNet~\cite{8460902} & 0.56 & 0.62 & 0.62 \\
\hline
\end{tabular}
\vspace{-0.3in}
\end{center}
\label{table:res}
\end{table}

\begin{table}[t]
\caption{Category-specific object affordance segmentation results of our model on the SOR3D-AFF test set based on video (top) and static image inference (bottom).}
\vspace{-0.1in}
\small
\setlength\tabcolsep{1.5pt} 
\renewcommand{\arraystretch}{0.9}
\begin{center}
\begin{tabular}{l c c c c c c c c c}
\hline
\\[-0.75em]
 \textbf{Metric} & cut & grasp & hammer & lift & paint & push & rotate & squeeze & type \\
\\[-0.85em]
\hline
\\[-0.75em]
IoU & 0.45 & 0.87 & 0.74 & 0.91 & 0.69 & 0.78 & 0.73 & 0.67 & 0.62 \\
F$_{\beta}$ & 0.57 & 0.92 & 0.85 & 0.94 & 0.77 & 0.87 & 0.84 & 0.74 & 0.71 \\
F$_{\beta}^{w}$ & 0.57 & 0.94 & 0.85 & 0.95 & 0.76 & 0.91 & 0.84 & 0.73 & 0.71 \\
\\[-0.75em]
\hdashline
\\[-0.75em]
IoU & 0.34 & 0.65 & 0.58 & 0.69 & 0.51 & 0.61 & 0.56 & 0.51 & 0.39 \\
F$_{\beta}$ & 0.41 & 0.68 & 0.62 & 0.73 & 0.54 & 0.65 & 0.61 & 0.55 & 0.45 \\
F$_{\beta}^{w}$ & 0.41 & 0.72 & 0.63 & 0.74 & 0.54 & 0.67 & 0.61 & 0.54 & 0.44 \\
\\[-0.85em]
\hline
\end{tabular}
\vspace{-0.3in}
\end{center}
\label{table:aff}
\end{table}

\begin{table}[t]
\caption{Comparative evaluation of a number of variations in the autoencoder architecture on the SOR3D-AFF test set based on video (top) and static image inference (bottom).}
\vspace{-0.1in}
\normalsize
\begin{center}
\begin{tabular}{l c c}
\hline
\\[-0.85em]
\textbf{Model Parameters} & IoU & F$_{\beta}$ \\
\hline
\\[-0.85em]
RGB & 0.62 & 0.73 \\
RGB + attention & 0.65 & 0.77 \\
RGB + attention + 2D flow & 0.66 & 0.78 \\
RGB-D & 0.64 & 0.77 \\
RGB-D + attention & 0.70 & 0.79 \\
RGB-D + attention + 3D flow & 0.72 & 0.80 \\
\\[-0.95em]
\hdashline
\\[-0.95em]
RGB + attention + 2D flow & 0.47 & 0.53 \\
RGB-D + attention + 3D flow & 0.54 & 0.58 \\
\hline
\end{tabular}
\vspace{-0.2in}
\end{center}
\label{table:abl}
\end{table}

\begin{figure}[t]
    \centering
    \includegraphics[width=0.9\columnwidth]{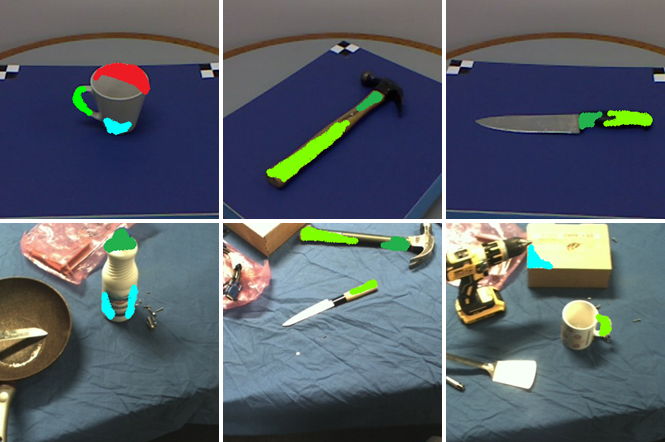}
    \vspace{-0.1in}
    \caption{Affordance label predictions on unseen objects from UMD~\cite{7139369} (top) and IIT-AFF~\cite{8206484} (bottom). The predictions are color-coded based on the SOR3D-AFF annotation: ``grasp" in light green, ``lift" in green, ``rotate" in red, ``push" in cyan.}
    \label{fig:qual}
    \vspace{-0.2in}
\end{figure}

\vspace{-0.1in}
\subsection{Qualitative Evaluation}
\label{qual}
Besides the quantitative evaluation on SOR3D-AFF, we choose to evaluate our model's ability to infer pixel-wise affordance labels on static unseen objects captured with different experimental settings. For this purpose, we use samples from the image-only UMD and IIT-AFF datasets. As depicted in Fig.~\ref{fig:qual}, our model is able to confidently predict learned affordances on similar objects of UMD, while also inferring reasonable labels of the dominant affordances on samples from the challenging, due to the cluttered scenes, IIT-AFF. Note that we center-crop a $300\times 300$ portion of each image and visualize the affordance label predictions with confidence greater than 0.75.

\vspace{-0.1in}
\section{Conclusion}
\label{sec:conclusion}
\vspace{-0.1in}
In this paper we proposed a novel approach for the task of object affordance segmentation. A deep autoencoder was presented, able to process RGB-D interaction sequences and implicitly localize the interaction hotspot through a soft-attention mechanism placed at the latent part of the model. Additionally, a corpus with RGB-D interaction videos coupled with pixel-wise annotation was introduced, as a subset of the SOR3D sensorimotor dataset. The presented experiments showed our model's competitive performance in comparison with the state-of-the-art, without the need of additional object-related information  such as bounding boxes and class labels. From the presented qualitative examples the model's ability to predict affordance labels on unseen objects was also demonstrated. Future work will investigate spatio-temporal affordance reasoning exploiting the capabilities of the presented attention mechanism. 
\bibliographystyle{IEEEbib}
\balance
\bibliography{main}

\end{document}